\documentclass[preprint]{elsarticle}

\usepackage{caption}
\usepackage{subcaption}
\usepackage{graphicx}
\usepackage{amsmath}
\usepackage{amssymb}
\usepackage{physics}
\usepackage{xcolor}
\usepackage{hyperref}
\usepackage{algorithm}
\usepackage{algpseudocode}
\usepackage[final]{showlabels} % Change [draft] here to [final] to turn off visible labels

\begin{document}
\begin{frontmatter}

\title{Tensor Network Estimation of Distribution Algorithms}

\author[1]{John Gardiner \corref{cor1}\fnref{fn1}}
\ead{john.gardiner@nasdaq.com}
\author[1]{Javier Lopez-Piqueres\fnref{fn1}}
\ead{jlopezpiquer@umass.edu}
\cortext[cor1]{Corresponding author }
\fntext[fn1]{Authors listed by alphabetical order.}

\affiliation[1]{
    organization={Quantum Computing Initiative, Nasdaq Inc.},
    addressline={11 Farnsworth St},
    city={Boston},
    state={MA},
    postcode={02210},
    country={USA}
}

\begin{abstract}
    Tensor networks are a tool first employed in the context of many-body quantum physics that now have a wide range of uses across the computational sciences, from numerical methods to machine learning.
    Methods integrating tensor networks into evolutionary optimization algorithms have appeared in the recent literature.
    In essence, these methods can be understood as replacing the traditional crossover operation of a genetic algorithm with a tensor network-based generative model.
    We investigate these methods from the point of view that they are Estimation of Distribution Algorithms (EDAs).
    We find that optimization performance of these methods is not related to the power of the generative model in a straightforward way.
    Generative models that are better (in the sense that they better model the distribution from which their training data is drawn) do not necessarily result in better performance of the optimization algorithm they form a part of.
    This raises the question of how best to incorporate powerful generative models into optimization routines.
    In light of this we find that adding an explicit mutation operator to the output of the generative model often improves optimization performance.
\end{abstract}


\end{frontmatter}

\section{Introduction}
\label{sec:intro}

There is a long history of interaction between machine learning models and evolutionary algorithms going at least as far back as the 1970s \cite{holland1977}.
This interaction has gone in both directions \cite{Banzhaf_2023}.
Evolutionary algorithms have been used to aid machine learning, for example, to tune hyperparameters of convolutional neural networks \cite{loshchilov2016cmaeshyperparameteroptimizationdeep}, search over model architectures \cite{miikkulainen2017evolvingdeepneuralnetworks}, or optimize parameters of neural networks in reinforcement learning algorithms \cite{salimans2017evolutionstrategiesscalablealternative}.
And conversely, machine learning models have been used to aid evolutionary algorithms, often by forming components of a larger evolutionary algorithm.
Examples abound: machine learning models have been used to define fitness functions \cite{jin2005comprehensive}, to define chromosome representations \cite{volz2018evolvingmariolevelslatent}, 
to perform ``smart" mutations \cite{openai2022}, and to generate new individuals from old by, in essence, performing a sophisticated form of crossover.
An early example of machine learning models used as a quasi-crossover component are Estimation of Distribution Algorithms (EDAs) \cite{muhlenbein1996recombination, bosman1999linkage, Larranaga_2002, hauschild2011survey, hu2012survey}.
EDAs are a family of optimization algorithms where ``parent" solutions are used to fit or update a generative model from which ``children" solutions are sampled.
Better solutions are then selected from among the children to become the next ``parents", i.e.\ the training data for the generative model of the next iteration. 
EDAs can be thought of as Genetic Algorithms (GAs) with traditional crossover (and mutation) replaced by a generative model. 
Algorithm \ref{alg: eda} describes the outline of an EDA in pseudocode. 
Step 6 consists of selecting $N$ individuals (parents) from the population $P$ using a chosen selection method. 
Common choices are greedy selection based on the fitness of top individuals, tournament selection, and Boltzmann selection where samples individuals are chosen proportionally to Boltzmann weights that depend on their fitness. 
Steps 7 and 8 correspond to learning and sampling from a probabilistic model. 
In step 7 the population of selected individuals $S$ is used in some way to obtain a new generative model $D$.
This step can involve fitting a model to the population $S$ \cite{larranaga2012estimation}, using $S$ to incrementally update the model from the previous generation \cite{baluja1994population}, or using $S$ as training data more generally.
Step 9 updates the population $P$ with the new samples $P'$.
This can involve replacing $P$ with $P'$, appending $P'$ to $P$, or combining them in some other way.

\begin{algorithm}
\caption{Estimation of Distribution Algorithm (EDA)}
\begin{algorithmic}[1]
\State \textbf{Input:} Number of parents $N$, number of children $M$, number of generations $G$
\State \textbf{Output:} Best solution found

\State Initialize population $P$ 
\For{$g = 1$ to $G$}
    \State Evaluate fitness of each individual in $P$
    \State Select $N$ individuals from $P$ to form the selected population $S$
    \State Learn a probabilistic model $D$ from $S$
    \State Sample $M$ new individuals from $D$ to form the new population $P'$
    \State Update population $P$ with $P'$
\EndFor
\State \Return the best individual found in $P$
\end{algorithmic}
\label{alg: eda}
\end{algorithm}

Using a generative model to produce new ``child" solutions in an evolutionary algorithm has a compelling conceptual basis.
Traditional genetic algorithms have well-documented difficulties optimizing many classes of functions.
For instance so-called deceptive functions \cite{deb1993analyzing}, functions with epistasis \cite{wright2004estimation} where fitness depends on correlations between entries far apart in a genetic encoding, or functions with Hamming cliffs \cite{zheng2007inquiry}
where nearness in the genetic encoding does not match nearness in the fitness landscape.
In the context of unsupervised generative models, generalization is the process by which a model infers a distribution from iid example data points \cite{zhao2018bias}.
Less formally, generalization is the task of generating new samples that are \emph{meaningfully} like given examples.
Ideally, crossover in a GA creates children that share meaningful beneficial features with their parents.
From this point of view the purpose of crossover is to perform a generalization task.
Traditional crossover is sometimes ill-suited to the task, however.
For example, traditional $k$-point crossover recombines blocks, contiguous entries in the genetic encoding.
This works well when the properties defining good solutions are encoded in blocks.
This is, of course, a strong assumption; in general, desirable properties may very well be encoded non-locally.
A more sophisticated procedure than traditional crossover would have the potential to learn salient features and how these features encode desirable properties and could thus generate children that are like the selected parent population in ways that are more consistently meaningful than simply sharing blocks.
EDAs instantiate this intuition by replacing crossover with a generative model, and there are indeed problem instances where EDAs provably perform better than GAs \cite{eda_vs_ga}.

\begin{figure}
    \centering
    \captionsetup{width=.9\linewidth}
    \includegraphics[width=0.9 \linewidth]{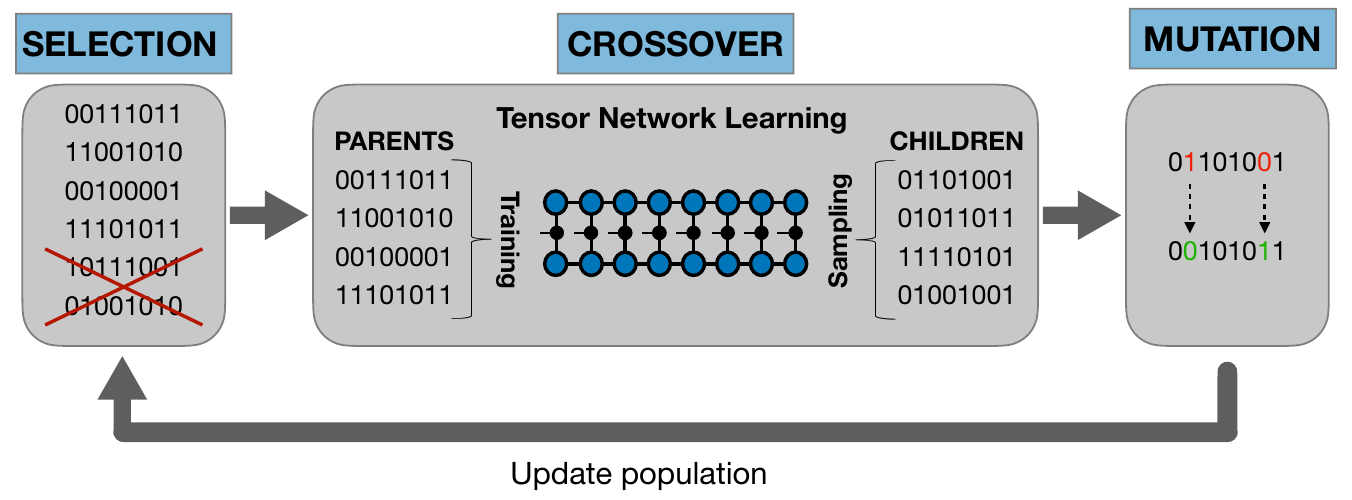} 
    \caption{\textbf{Tensor Network EDA.} First a selection procedure screens top bit string candidates from a pool (the \textit{population}). Next, a Tensor Network Generative Model (shown an example of a Born Machine) trains on those candidates (the \textit{parents}), and outputs new ones resembling the original (the \textit{children}). Finally, a mutation operator flips at random output samples. New and unique samples are then added back to the pool. This series of steps is repeated throughout many iterations.}
    \label{fig:tn-eda}
\end{figure}

The probabilistic model at the heart of an EDA can be any kind of generative model from the very simple, like independent Bernoulli distributions \cite{umda}, to the sophisticated, like GANs \cite{probst2015generative}.
Of interest to us are tensor-network based generative models. 
Tensor networks (TNs) are a tool for representing multi-dimensional arrays as a composition of other, often simpler or smaller, arrays.
They are a tool first employed in the context of many-body quantum physics that now has a wide range of uses, including generative modeling \cite{han2018unsupervised, cheng2019tree}.
Advantages of tensor-network based generative models include efficient sampling \cite{perfect_sampling} as well as greater interpretability compared to some alternatives \cite{ran2023tensor}.
Two recent works propose evolutionary algorithms with tensor network generative model components.
In \cite{original_geo} the authors explore an algorithm they call tensor network generator enhanced optimization (GEO), and in \cite{batsheva2024protes} the authors explore an algorithm they call probabilistic optimization with tensor sampling (PROTES).
These algorithms have their differences, including in the particular tensor network generative model employed, but they both fit within the EDA framework of iteratively modeling a selected population, then sampling from the model to obtain new, hopefully better, solutions for the population.
We describe them in more detail in Section \ref{sec:background}. 
Such tensor network EDAs can be extended to incorporate an explicit mutation step. 
While a mutation step is not a definitional part of the EDA framework \cite{larranaga2012estimation}, previous work has suggested the effectiveness of adding an explicit mutation step to keep diversity of samples through generations \cite{handa2007effectiveness}. 
In Fig. \ref{fig:tn-eda} we show the pipeline of a tensor network EDA, including an explicit mutation step. 

\subsection{Contributions} \label{contr}
Our contributions can be summarized as follows:
\begin{itemize}
    \item We identify two recent proposals using tensor networks \cite{original_geo, batsheva2024protes} as probabilistic models within an evolutionary algorithm for solving combinatorial optimization problems as subclasses of a much older framework: that of Estimation of Distribution Algorithms (EDAs). 
    \item We explore the function of the tensor-network generative model. 
    Our results show that a better (at generalization) generative model does not necessarily lead to a better EDA optimizer. 
    In fact, we consistently see that adding noise to the generative model in the form of an explicit mutation step after sampling leads to better optimization performance.
    We recommend that practitioners add such a mutation step because of its simplicity and effectiveness.
    \item We compare different variations of EDA with different selection procedures, with different probabilistic models, and with/without mutation. 
    Our findings suggest that a low-expressivity probabilistic model, in our case a TN corresponding to a matrix product state of bond dimension 2 and a Markov chain Bayesian network, with a selection procedure based on a Boltzmann distribution of historical records, together with a mutation operator applied to sampled data from the model, results in effective optimizers.  
\end{itemize}

\section{Background}
\label{sec:background}
\subsection{Tensor Networks} 
\label{sec:tensor_nets}
A tensor network is a set of multi-dimensional arrays where some pairs of indices are summed over \cite{schollwock2011density, orus2014practical, cichocki2018tensor, banuls2023tensor}.
They can be seen as a generalization of the idea of matrix decomposition to higher-dimensional arrays.
For instance, consider the N-dimensional tensor with entries $f(x_1, x_2, \cdots, x_N)$ with $x_i \in \{0,1\}$ and $f$ some scalar function on $\{0, 1\}^N$. 
One popular decomposition of such a tensor is the matrix product state (MPS) tensor network with form 
\begin{equation}
    f(x_1,x_2,\cdots, x_N)=\sum_{l_1,l_2,\cdots, l_{N-1}}T_{x_1}^{[1]l_1}T_{x_2}^{[2]l_1 l_2}\cdots T_{x_N}^{[N]l_{N-1}}.
\end{equation}
Each tensor $T^{[i]}_{x_i}$ is a matrix with row/column indices $(l_i,l_{i+1})$, except for the first/last one which correspond to a row/column vector. 
Here the indices $l_i=1,2,\cdots, \chi_i$ denote the \textit{link} or \textit{bond} indices of each of the pair connected tensors. 
The maximum value of each bond index $\chi_i$ is known as the rank or bond dimension of that bond. 
An MPS tensor network decomposition where the bonds have dimension $\chi$ can be represented with $\mathcal{O}(N\chi^2)$ values while the original tensor would require $2^N$.
This linear, versus exponential, scaling in $N$ makes the MPS a useful tool for efficiently compressing high dimensional arrays.

When dealing with tensor networks it is often useful to use a pictorial representation where tensors are represented by shapes (here circles), indices are lines leaving the shapes, and contractions are represented by joining these lines.
The pictorial representation of an MPS is quite simple: 
\[\includegraphics[height=14ex]{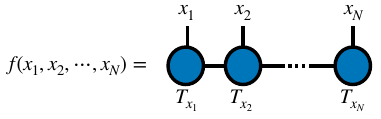} \]
Pictorial representations become especially helpful when dealing with more nontrivial decompositions. 
Tensor network factorizations can sometimes be constructed via a series of suitable matrix decompositions, such as singular value decomposition. 
The power of tensor networks is in the realization that some high-order tensors afford an efficient representation in terms of low-order tensors of \textit{low rank}, where the bond dimension scales polynomially $\chi\sim \mathcal{O}(\text{poly}(N))$ \cite{cichocki2014tensornetworksbigdata, cichocki2016tensor, cichocki2017tensor}.

\subsection{Tensor Network Born Machines and Bayesian Networks}
\label{sec:born_bayes}
\begin{figure}
    \centering
    \captionsetup{width=.9\linewidth}
    \includegraphics[width=0.8\linewidth]{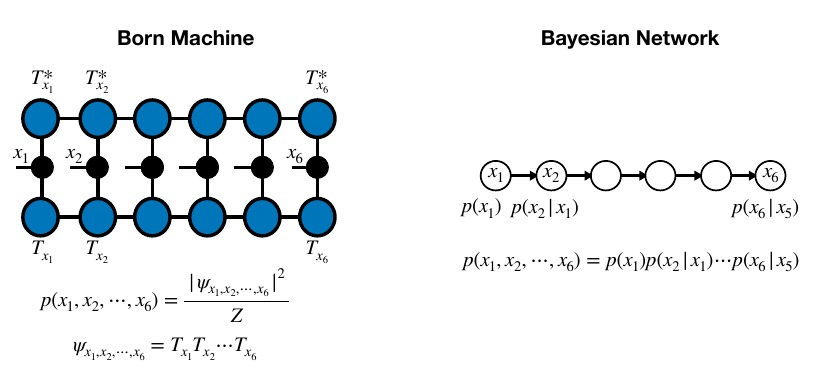}
    \caption{
        \textbf{Born Machine and Bayesian Network}. 
        \textit{(Left):} 
        Born Machines are inspired by the probabilistic interpretation of quantum mechanics. 
        Shown an example of a Matrix Product State Born Machine where probabilities are expressed as the complex conjugate square of amplitudes, which in turn are described by an MPS. 
        \textit{(Right):}
        Bayesian Networks are directed acyclic graphs where nodes represent random variables and edges represent conditional dependencies.
        Shown an example of a Bayesian Network with chain topology, i.e.\ a Markov Chain.
    }
    \label{fig:Born_vs_Bayes}
\end{figure}

One useful application of tensor networks is in modeling high-dimensional probability distributions.  
Some tensor network decompositions like MPS allow not only for efficient representations of probability distributions as well as its marginals, but equally important, an efficient and exact sampling technique known as \textit{perfect sampling} \cite{perfect_sampling}. 
There are two commonly encountered ways of modeling a probability distribution with a tensor network.
The first is to represent the probabilities directly.
The tensors are constrained to have non-negative entries and the value of the tensor network is understood as an unnormalized probability.
For an MPS, for example, this would read $p(x_1,x_2,\cdots, x_N)=T^{[1]}_{x_1}T^{[2]}_{x_2}\cdots T^{[N]}_{x_N}/Z$ with $Z=\sum_{x_1,x_2,\cdots,x_N}T^{[1]}_{x_1}T^{[2]}_{x_2}\cdots T^{[N]}_{x_N}$, with $T^{[i]}_{x_i}\geq 0$. 
The second way, inspired by quantum mechanics, is to encode amplitudes, quantities whose absolute values squared are probabilities, rather than the probabilities themselves.
Such a tensor network model is known as a Tensor Network Born Machine (TNBM). 
For an MPS this would be $p(x_1,x_2,\cdots, x_N)=|\psi_{x_1,x_2,\cdots, x_N}|^2/Z$, with $\psi_{x_1,x_2,\cdots,x_N}=T^{[1]}_{x_1}T^{[2]}_{x_2}\cdots T^{[N]}_{x_N}$ and $Z=\sum_{x_1,x_2,\cdots,x_N}|\psi_{x_1,x_2,\cdots, x_N}|^2$. While the TNBM ansatz allows for complex amplitudes, we only consider real amplitudes in this work. 
Note that although the sums defining $Z$ in both MPS models involve exponentially many terms, they can in fact be performed efficiently via tensor network contraction.
For example, in the case where the MPS models probabilities directly,
\begin{equation*}
Z=\sum_{x_1,x_2,\cdots,x_N}T^{[1]}_{x_1}T^{[2]}_{x_2}\cdots T^{[N]}_{x_N}
=\left(\sum_{x_1}T^{[1]}_{x_1}\right)\left(\sum_{x_2}T^{[2]}_{x_2}\right)\cdots \left(\sum_{x_N}T^{[N]}_{x_N}\right)
\end{equation*}
reducing the exponential sum to the multiplication of $\mathcal{O}(N)$ matrices.

The graph structure of the tensor network ansatz for modeling probability distributions resembles that of a probabilistic graphical model (PGM). 
The connection between PGMs and TNs has been explored in the past \cite{robeva2017dualitygraphicalmodelstensor, glasser2019expressive, srinivasan2020quantumtensornetworksstochastic, miller2021probabilisticgraphicalmodelstensor}. 
PGMs are frequently employed as generative models within EDAs due to their capacity to represent and factorize distributions as products of marginal probability distributions over correlated variables \cite{hauschild2011survey}. 
This factorization property, which is central to the tensor network's probabilistic model-building capability, facilitates efficient construction and sampling processes. 
As a result, PGMs have become a preferred model choice in EDAs. 
In Fig.~\ref{fig:Born_vs_Bayes} we show an MPS Born machine and a one-dimensional Bayesian network (BN), a type of PGM widely used in EDAs.  

% \subsection{Training Tensor Network Born Machines} \label{tnbm_training}
There are a variety of ways to train tensor network generative models (or tensor network models more generally) including gradient descent \cite{han2018unsupervised, novikov2021tensor}, tensor cross interpolation \cite{tensor_cross_interpolation}, and tensor sketching \cite{tensor_train_sketching}.
The experiments in this work use the gradient descent method of \cite{han2018unsupervised} a description of which is presented in \ref{appendix:training_algo}.

\subsection{Generator-enhanced Optimization (GEO)}
\label{sec:geo_intro}

\begin{figure}
    \centering
    \captionsetup{width=.9\linewidth}
    \includegraphics[width=0.9\linewidth]{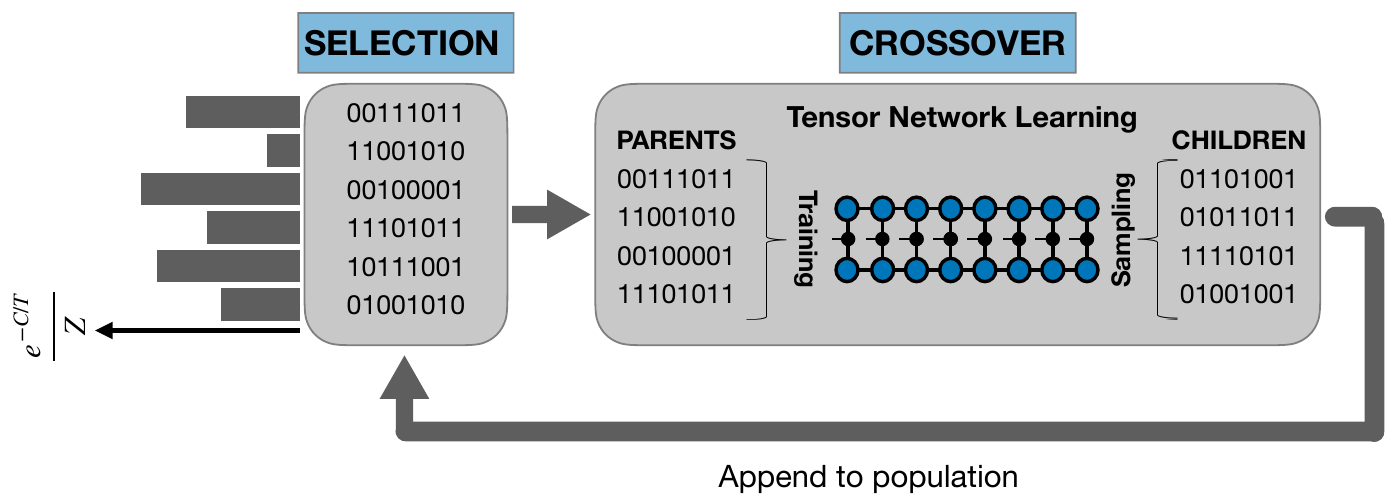}
    \caption{\textbf{GEO pipeline \cite{original_geo}.} A Tensor Network (MPS in the original work) Born Machine generative model is trained to produce high quality samples by constructing a population of samples from the model. 
    At each iteration a Boltzmann selection procedure selects training samples based on quality from all samples from all previous iterations. 
    The tensor network can be swapped out for a different generative model.}
    \label{fig:geo_pipeline}
\end{figure}

In \cite{original_geo} the authors introduce a framework, which they call generator-enhanced optimization (GEO), for incorporating generative models into an evolutionary algorithm.
In the resulting algorithm, solutions to an optimization problem are kept in a pool.
At each iteration high-quality solutions in the pool are selected and used to train a generative model which is then sampled to get new, potentially higher-quality solutions.
The new solutions are added to the pool if they satisfy the optimization problem constraints.
Because at each iteration the generative model is trained on solutions selected for their quality, the hope is that new sampled solutions will be of comparable, and occasionally better, quality.
Over several iterations the quality of the best solutions increases. 
See Fig. \ref{fig:geo_pipeline}. 

The generative model used by \cite{original_geo} and follow-up work \cite{generalization_metrics_geo, bmw_geo, symmetric_geo, linearly_constrained_geo} is a TNBM with an MPS tensor network. 
To train the TNBM on the selected data points, one can minimize the training negative log-likelihood via gradient descent, for example, using the procedure described in \ref{appendix:training_algo}.

Using TNBMs in their GEO framework the authors of \cite{original_geo} obtain state of the art results on certain financial portfolio optimization problems and in later work show that it can be used with some success on optimization problems with black-box functions that are expensive to compute \cite{bmw_geo}.

In the GEO algorithm the selection of the training data for the model is done by drawing iid points from an explicitly given distribution.
In our examples and those of \cite{original_geo} this is a Boltzmann distribution $p(x)\sim e^{-f(x)/T}$ where $f(x)$ is the objective function to be minimized and $T$ is a temperature that can vary from iteration to iteration, either dynamically or according to a fixed schedule.
Because the training data are iid draws from a known distribution, we can precisely assess how well models perform on the generalization task of outputting samples from the distribution their training data is taken from.
We make use of this in Section \ref{sec:worse_models} to compare models.

\subsection{Probabilistic Optimization with Tensor Sampling (PROTES)}
\label{sec:protes}
\begin{figure}
    \centering
    \captionsetup{width=.9\linewidth}
    \includegraphics[width=0.8\linewidth]{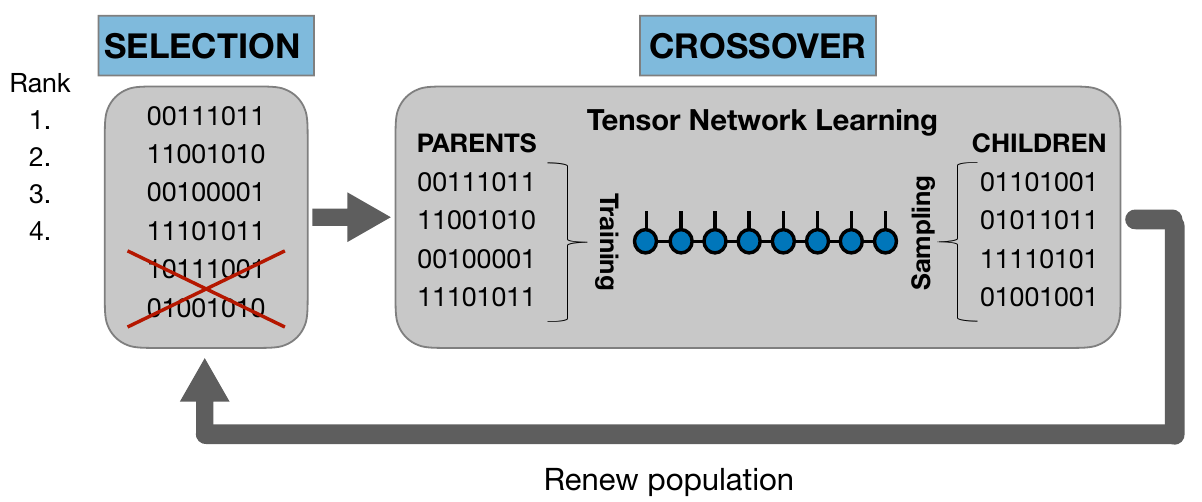}
    \caption{\textbf{PROTES pipeline \cite{batsheva2024protes}.} At each iteration, a positive Tensor Network (an MPS in the original work) is trained to produce high quality samples by constructing a pool of samples from the model from the previous iteration. 
    Only the top candidates are selected from the pool as training data for the next iteration.}
    \label{fig:protes_pipeline}
\end{figure}
A different evolutionary algorithm involving a tensor network was recently proposed by \cite{batsheva2024protes} under the name PROTES. 
The generative model is in this case a positive MPS, where the tensor network encodes probabilities directly rather than amplitudes as in a Born machine.
The way PROTES solves the combinatorial optimization problem at hand is very similar to GEO, with one difference being that the candidates to train over are chosen greedily and are chosen only from the most recent generation of samples from the model (not from a growing bank of solutions).
Further, the model distribution in PROTES is incrementally updated generation to generation, while in GEO the training procedure of the model distribution at a given generation is let unspecified (in particular, one is free to swap the model used in the previous generation by a new one). 
An illustration of the PROTES algorithm is shown in Fig. \ref{fig:protes_pipeline}. 
PROTES was found to beat several popular heuristic algorithms in 19 out of 20 selected optimization problem instances \cite{batsheva2024protes}.

Gradient descent with negative log-likelihood forms a part of both PROTES and GEO as implemented by \cite{original_geo}, but with somewhat different interpretations.
In GEO, optimizing the log-likelihood is framed within a generative modeling approach, where the goal is to align the model distribution with a target distribution by minimizing the KL-divergence. 
In contrast, the probabilistic model in PROTES is more akin to a reinforcement learning policy function, and the log-likelihood optimization arises when using the REINFORCE trick \cite{williams1992simple}. 

\section{Better Optimization Using Worse Generative Models}
\label{sec:worse_models}
Many stochastic optimization algorithms can be understood in terms of an exploit/explore dichotomy.
The algorithm uses information gained from previously evaluated solutions to suggest new solutions that are promising but also suggests solutions with an eye on collecting more information.
The algorithm performs well when sampling within regions already identified as promising is balanced with exploring new regions.

The GEO algorithm, for example, is subject to these considerations of balancing exploration and exploitation.
The TNBM generative model at the heart of GEO exploits knowledge of already evaluated solutions to generate solutions of good quality.
Exploration can be provided by the inductive bias of the model which will keep its output from perfectly modeling the distribution from which training data is drawn.
Exploration might also be provided, for example, by the model being under trained and retaining some of the randomness of its initial parameter values.
In any case, generated solutions are typically new solutions not previously seen.
These new solutions may very well be better by chance and lead the algorithm to more promising regions of solution space.

One would imagine that GEO works well only to the extent that these channels of exploration and exploitation are balanced appropriately.
Whether this is the case will depend in part on the choice of generative model and selection method.
But unlike genetic algorithms, say, which have a mutation step, GEO as originally conceived has no step explicitly dedicated to exploration separate from exploitation.
This can make it difficult to tune GEO performance in practice.
This can also lead to the perverse situation wherein a generative model can be \emph{better} along typical metrics for generative model performance but lead to a \emph{worse} GEO optimizer.

Indeed, we find in experiments that adding arbitrary noise to the generative model sampling procedure can improve the performance of GEO.
This noise can be in the form of random bit flips occurring after the sampling process or as Gaussian noise added to the tensor entries of the tensor network.
This noise, while improving the performance of GEO, serves to increase the KL-divergence between the model distribution and the distribution from which training data is sampled, suggesting that the noise makes the model a worse generative model as typically understood.
We also find that lowering the expressiveness of the TNBM, as measured by the bond dimension of the tensor network, can improve GEO performance even in a regime where it increases the KL-divergence relative to the underlying distribution.
We report the results of these experiments in this section.
Details of the experiments, including GEO parameters as well as model training details are provided in \ref{appendix:port_opt_details}.

Based on these results, we suggest an alternative to GEO as originally formulated wherein a generative model is used as a crossover step in a genetic algorithm that includes an explicit mutation step, in other words a tensor-network EDA with mutation.
This gives the user a tunable parameter (the mutation rate) with which to balance exploration against the exploitation inherent in the generative model.
Note that the point of the mutation operator is \emph{not} as a regularizer on the generative model to avoid overfitting.
Indeed, in our experiments, we see that the mutation operator \emph{raises} the KL-divergence relative to the distribution from which training data is taken.

The addition of an explicit, tunable exploration step may allow for the successful use of a wider variety of generative models, including more expressive models.
It helps to remove the concern that improving the generative model used might lead to a worse optimizer.

\subsection{Equal-weighted Portfolio Optimization}\label{sec:port_opt}
For our first experiments we take equal-weighted portfolio volatility minimization as our benchmark problem.
An equal-weighted portfolio is one where each asset present in the portfolio constitutes an equal fraction of the total by value.
The task is to choose assets to include or not in the portfolio so that the variance of the returns is minimized, subject to the constraint that all assets included in the portfolio have the same weight.
We can further include the constraint that the number of assets included in the portfolio be within some range.

Let $\Sigma$ denote the covariance matrix of asset returns, and let $x$ be a vector of binary digits representing the inclusion or exclusion of each asset.
For an equal-weighted portfolio this implies asset weights $w=x/(x^Tx)$.
The variance portfolio returns (volatility squared) of the portfolio is $w^T\Sigma w$.
So the objective function to be minimized can be written
\begin{equation*}
f(x) = 
\begin{cases}
\frac{x^T\Sigma x}{\left(x^Tx\right)^2} & \text{if $n_\text{min} \leq x^Tx \leq n_\text{max}$,}\\
\infty & \text{else},
\end{cases}
\end{equation*}
where $n_\text{min}$ and $n_\text{max}$ are the aforementioned cardinality bounds, the minimum and maximum number of assets allowed in the portfolio.
For reasons of practicality, one can also make the cardinality constraint soft by instead returning $f(x) = \left(x^Tx - n_\text{max}\right)C$ or $f(x) = \left(n_\text{min} - x^Tx\right)C$ in the case where the cardinality of $x$ is above or below the bounds respectively, where $C$ is some sufficiently large constant.

% \subsubsection{Noisy Generative Models}\label{noisy_models}

In our experiments with this portfolio optimization problem,
we explored three ways of degrading the quality of the generative model at the heart of the GEO protocol. 
First, we consider noise in the form of independent bit flips, with a probability $p_{\mathrm{flip}}$, for each bit in a sampled output of the model.
Second, we consider the addition of independent Gaussian noise to every entry of every tensor in the tensor network.
Third we compare models with different expressivity, by reducing the bond dimension of the tensor network.

In all three cases we find that the noise or lowered expressivity increases the KL-divergence between the outputs of the model and the distribution from which the training data is sampled, confirming that these modifications are indeed degradations of the model.
Crucially, we find in the experiments that these model degradations can improve the performance of GEO as an optimizer.

% We find in both cases that increasing noise increases the KL-divergence between the outputs of the model and the distribution from which the training data is sampled, but also that adding noise can improve the performance of GEO as an optimizer.

\subsubsection{Bit-flip Noise}
\label{sec:bit_flip_noise}

\begin{figure}
    \centering
    \captionsetup{width=.9\linewidth}
    \begin{subfigure}[t]{0.47\textwidth}
        \centering
        \includegraphics[width=\linewidth]{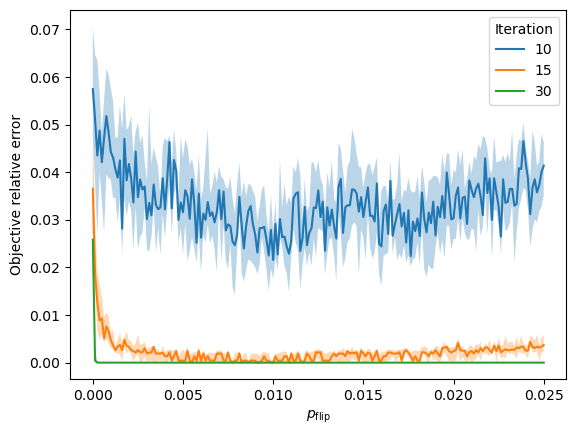} 
    \end{subfigure}
    \hspace{0.02\textwidth} 
    \begin{subfigure}[t]{0.47\textwidth}
        \centering
        \includegraphics[width=\linewidth]{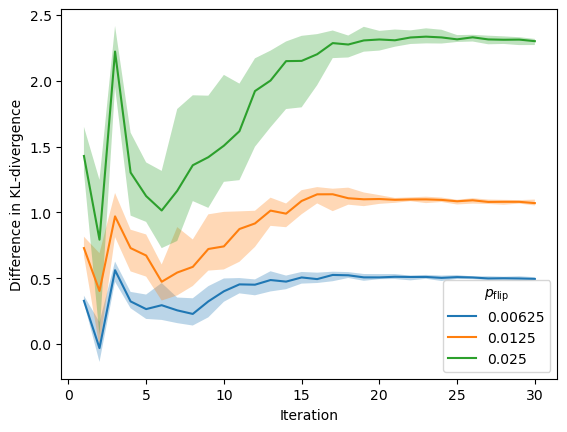} 
    \end{subfigure}
    \begin{subfigure}[t]{0.47\textwidth}
        \centering
        \includegraphics[width=\linewidth]{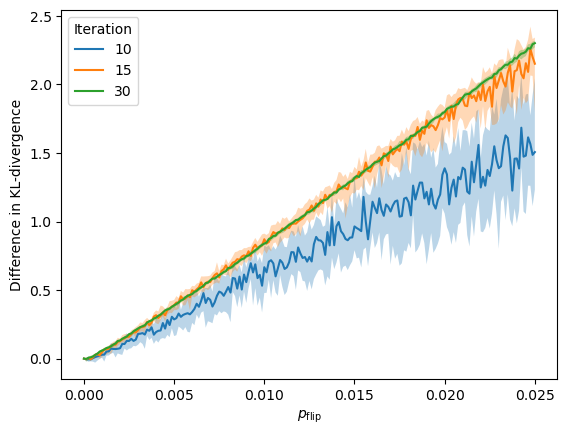} 
    \end{subfigure}
    \caption{
        \textbf{Portfolio Optimization: GEO with bit-flip noise} 
        \textit{(Top Left):} GEO performance (on the equal-weighted portfolio optimization problem) with various values of $p_\text{flip}$.
        Lower is better.
        Lines are median values out of 40 independent runs.
        Shaded regions are 1st and 3rd quartiles. 
        \textit{(Top Right and Bottom):} At each iteration a noiseless ($p_\text{flip}=0$) generative model is trained in parallel to the model used for GEO.
        Plotted are the KL-divergence for the noisy model minus the KL-divergence for the noiseless model.
        These KL-divergences are relative to the Boltzmann distribution from which the training data that iteration is drawn.
        Lines are medians and shaded regions are between 1st and 3rd quartiles out of 40 independent runs.
        We see that KL-divergence is nearly always higher for the noisy model and that the difference in KL-divergence between noisy and noiseless models increases with increasing noise (increasing $p_\text{flip}$).
    }
    \label{fig:bit_flip_noise}
\end{figure}

We ran GEO for the equal-weight portfolio optimization problem described above in \ref{sec:port_opt}.
Solutions to this problem are asset portfolios represented by bit strings where each entry denotes whether a given asset is absent (0) or present (1) in the portfolio.
In each iteration of GEO a generative model is trained on selected solutions to the optimization problem, and new solutions are sampled from the generative model.
In this experiment, we modified the output of the generative model by independently flipping each bit in the portfolio bit string with a probability $p_\text{flip}$.
Composing the original generative model together with this bit-flip mutation constitutes a new generative model that is effectively a ``noisy" or diffused version of the original.
To the extent that the generative model has learned the underlying distribution from which its training data is drawn, this diffusion should result in a higher KL-divergence relative to this distribution.

We ran GEO with this bit flip modification for values of $p_\text{flip}$ from 0 to 0.025.
For details of the GEO parameters used see \ref{appendix:port_opt_details}.
Performance is plotted in the top left of Fig.~\ref{fig:bit_flip_noise}, where the objective function value relative error is computed w.r.t. the optimum one found using \textsc{Mosek} \cite{mosek}.
Median performance was best for bit-flip rates near $p_\text{flip}=0.01$.
This significantly outperformed GEO without the additional bit-flip mutations, the noiseless case with $p_\text{flip}=0$, which fails to find the optimal solution even after 30 iterations of GEO.

For comparison, at each iteration of the GEO algorithm the original, un-bit-flipped generative model is trained in parallel on the same training data as the noisy, bit-flipped model used in the algorithm.
As expected, we find that in general the generative model composed with a bit-flip mutation step has higher KL-divergence after training than the original generative model alone.
This is true for nearly every run and nearly every iteration.
See the top right of Fig.~\ref{fig:bit_flip_noise}.
The difference in KL-divergence between the noisy and noiseless models also increases with increasing $p_\text{flip}$, again as expected.
See the bottom plot of Fig.~\ref{fig:bit_flip_noise}.
(See \ref{appendix:kl_divergence} for an explanation of how KL-divergence in the presence of bit-flip noise can be efficiently calculated exactly for the tensor network generative model.)
We see a pattern wherein increasingly worse generative models lead, at least at first, to \emph{better} performance of the GEO framework.

We emphasize that these generative models are indeed worse.
They are worse at the job of modeling the distribution from which their training data is taken, which is to say that these generative models are worse \emph{qua} generative models.
This adds nuance to an understanding of GEO wherein the generative model's purpose is understood to be learning features of good solutions and generalizing to unseen good solutions, as this is \emph{the} task that better generative models can do better.
Beyond exploiting features of already observed solutions, the generative model in GEO evidently provides other functions.
In particular we hypothesize it also provides an exploration function, a function that at least in some cases a strictly worse generative model can fulfill better.
As we will discuss in Section \ref{sec:discussion} this suggests that an alternative set up that explicitly separates out these two functions, so that performance on one is no longer in trade-off with performance on the other, might perform better than GEO as originally conceived.

\subsubsection{Noisy Tensor Entries}
\label{sec:noisy_entries}

\begin{figure}
    \centering
    \captionsetup{width=.9\linewidth}
    \begin{subfigure}[t]{0.47\textwidth}
        \centering
        \includegraphics[width=\linewidth]{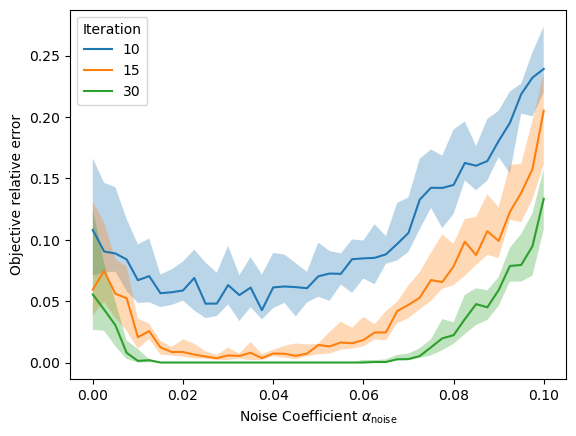} 
    \end{subfigure}
    \hspace{0.02\textwidth} 
    \begin{subfigure}[t]{0.47\textwidth}
        \centering
        \includegraphics[width=\linewidth]{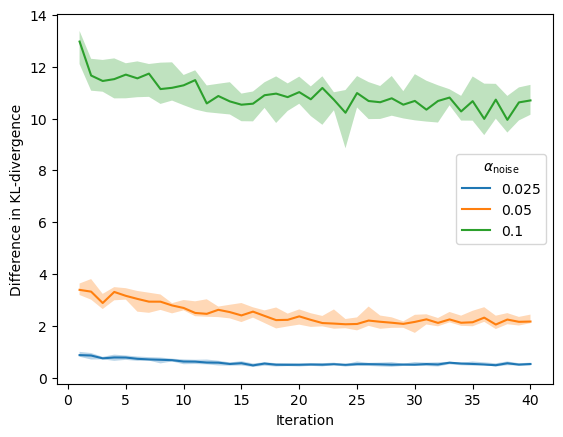} 
    \end{subfigure}
    \begin{subfigure}[t]{0.47\textwidth}
        \centering
        \includegraphics[width=\linewidth]{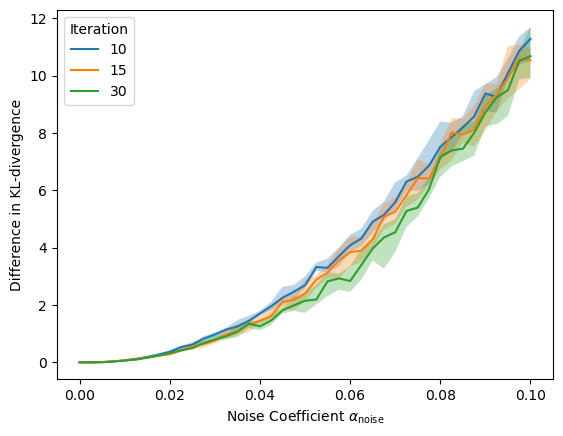} 
    \end{subfigure}
    \caption{
        \textbf{Portfolio Optimization: GEO with noisy tensor entries} 
        \textit{(Top Left):} GEO performance with tensor entries corrupted by Gaussian noise with standard deviation $\alpha_\text{noise}$.
        Lower is better.
        Lines are medians and shaded regions are 1st and 3rd quartiles out of 40 runs.
        \textit{(Top Right and Bottom):} 
        Difference in KL-divergence between the noisy and noiseless ($\alpha_\text{noise}=0$) generative models.
        At each iteration a noiseless ($\alpha_\text{noise}=0$) generative model is trained on the same data as the noisy model used for GEO.
        KL-divergence is relative to the Boltzmann distribution from which the training data that iteration is drawn.
        Lines are medians and shaded regions are 1st and 3rd quartiles out of 40 independent runs.
        We see that KL-divergence is higher for the noisy model and that the difference in KL-divergence between noisy and noiseless models increases with increasing noise (increasing $\alpha_\text{noise}$).
    }
    \label{fig:noisy_entries}
\end{figure}

We ran a second experiment where at each iteration, after training a tensor network generative model, the entries of every tensor were independently modified with noise.
Specifically, to each entry of each tensor was added an independent value drawn from a normal distribution $\mathcal{N}(0, \alpha_\text{noise})$.
Unlike the bit flips of Section \ref{sec:bit_flip_noise}, this noise in some sense reflects the internal structure of the tensor network model we use.

We recorded the performance of GEO on the equal-weighted portfolio optimization problem with several values for the noise coefficient $\alpha_\text{noise}$.
We find that the optimal amount of noise is not zero.
Rather, setting the noise coefficient near $\alpha_\text{noise}=0.035$ resulted in the best performance.
See the top left plot of Fig.~\ref{fig:noisy_entries}.
Adding noise to the tensor entries makes the generative models worse in the sense that it increases its KL-divergence relative to the distribution from which its training data is drawn.
We confirmed that this was indeed the case for the models used in GEO by calculating the KL-divergence of the trained generative model before and after the addition of noise and comparing the values.
Furthermore we found that in the context of a GEO run the generative models in runs with a higher $\alpha_\text{noise}$ have a higher difference in KL-divergence from the noise-free model.
See the top right and bottom plots of Fig.~\ref{fig:noisy_entries}.
All this is to confirm that the generative models in this experiment that lead to the best performance of GEO are not the models that are best at the core task of any generative model: modeling the distribution from which their training data is drawn.

% \subsubsection{Low Expressivity Models}\label{low_expressivity_models}
\subsubsection{Low Bond-dimension MPS Models}
\label{sec:low_bond_dimension}
In Sections \ref{sec:bit_flip_noise} and \ref{sec:noisy_entries} we show that generative models degraded by the presence of noise can lead to better optimization performance when used in the GEO framework.
Here we consider a different way of degrading a model: making it less expressive.
In the context of the TNBM model we use, expressivity is influenced by the rank (or ``bond dimension") of the tensor network.
This is the dimension of the indices that are contracted between neighboring tensors in the network.
Higher bond dimensions allows for modeling correlations between features that are farther apart in the network as well as modeling more complex correlations for nearby features.

\begin{figure}
    \centering
    \captionsetup{width=.9\linewidth}
    \begin{subfigure}[t]{0.47\textwidth}
        \centering
        \includegraphics[width=\linewidth]{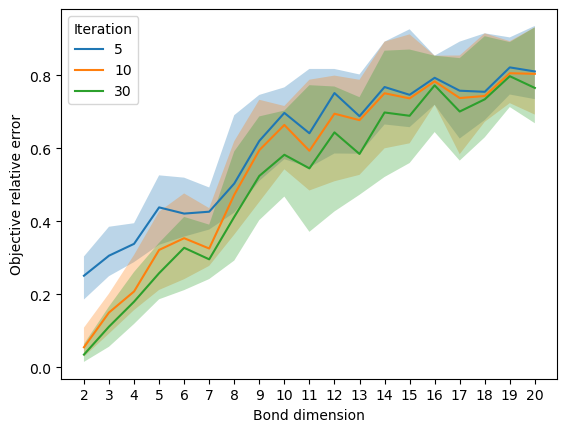} 
    \end{subfigure}
    \hspace{0.02\textwidth} 
    \begin{subfigure}[t]{0.47\textwidth}
        \centering
        \includegraphics[width=\linewidth]{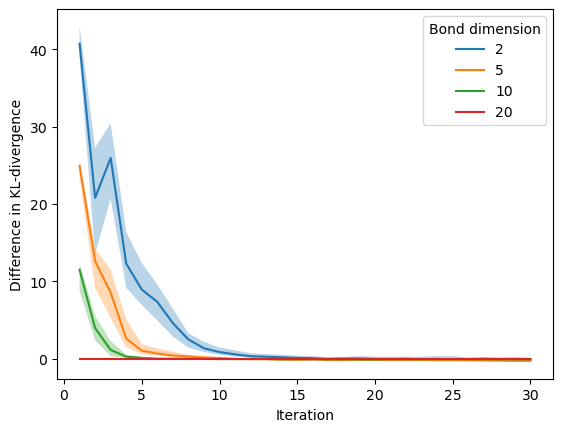} 
    \end{subfigure}
    \begin{subfigure}[t]{0.47\textwidth}
        \centering
        \includegraphics[width=\linewidth]{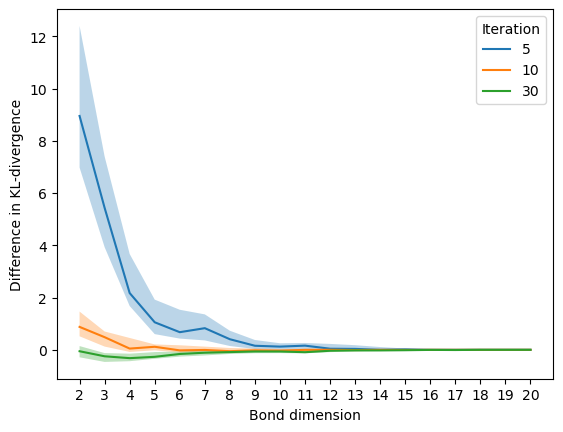} 
    \end{subfigure}
    \caption{
        \textbf{Portfolio Optimization: GEO using tensor networks of various bond dimensions} 
        \textit{(Top Left):} Performance (lower objective is better) of GEO using an MPS tensor network with bond dimensions ranging from 2 to 20.
        Line is median and fill is between 1st and 3rd quartiles out of 40 independent runs.
        We see performance tends to decrease with increasing bond dimension and hence a more expressive model.
        \textit{(Top Right and Bottom):} Difference in KL-divergence between the tensor network used as part of GEO and a tensor network with bond dimension 20 trained on the same data.
        The KL-divergence is assessed every iteration after training.
        We see that KL-divergence for the models with bond dimension 2-19 is generally higher than for the bond-dimension 20 model.
        That is to say the bond-dimension 20 model better learns the distributions from which training data is drawn.
    }
    \label{fig:low_bonddim}
\end{figure}
We ran GEO with different values for maximum bond dimension of the TNBM model.
At each iteration of GEO, in addition to the generative model with the chosen bond dimension, we also train a reference generative model with higher bond dimension (bond dimension 20) on the same training data.
We find that the less expressive (lower bond dimension) models have a higher KL-divergence in general, and hence are worse models as measured by their ability to model the distribution from which their training data is sampled, but that a GEO run with higher bond dimension generative models often performs worse.
See Fig.~\ref{fig:low_bonddim}.

In \cite{bmw_geo} the authors describe problem instances where a bond dimension of 6 gives better GEO performance than higher bond dimensions.
Those authors suggest that the observed worse performance of higher bond dimensions in their GEO runs is related to overfitting. 
Indeed, it is in general easier to overfit with a higher bond dimension, so for a fixed training protocol it is certainly possible for a higher-bond dimension model to overfit and thereby do a worse job modeling the distribution from which training data is drawn.
The above experiment, however, suggests that no such explanation in terms of overfitting is necessary: more expressive models can lead to worse GEO performance even when overfitting is not present, as evidenced by no increase in KL-divergence for the more expressive models.

\subsection{Neural Architecture Search}
\label{sec:nas_noisy_entries}
Beyond portfolio optimization, we consider Neural Architecture Search (NAS).
% as a second application domain for TN-EDA. 
NAS can be formulated as a discrete black-box optimization problem where each candidate neural network architecture is defined by a structured set of categorical decisions (e.g., operation types and connectivity choices) and where the objective value is validation performance after training. 
This setting is a natural fit for model-based EDAs: the search space is combinatorial, variables are strongly dependent, and evaluations are expensive enough that sample efficiency is critical.

We choose the \textsc{NAS-Bench-301} benchmark from \cite{zela2022surrogatenasbenchmarksgoing} over other benchmark problems because it provides a surrogate-based evaluation protocol while retaining a substantially larger and more realistic search space than earlier tabular NAS benchmarks. 
The domain of \textsc{NAS-Bench-301} is the DARTS \cite{liu2019dartsdifferentiablearchitecturesearch} search space of convolutional neural network cells with over $10^{18}$ distinct architectures.
The pre-trained surrogate evaluation model of \textsc{NAS-Bench-301} predicts performance of any such architecture on CIFAR-10.

\begin{figure}
    \centering
    \captionsetup{width=.9\linewidth}
    \begin{subfigure}[t]{0.47\textwidth}
        \centering
        \includegraphics[width=\linewidth]{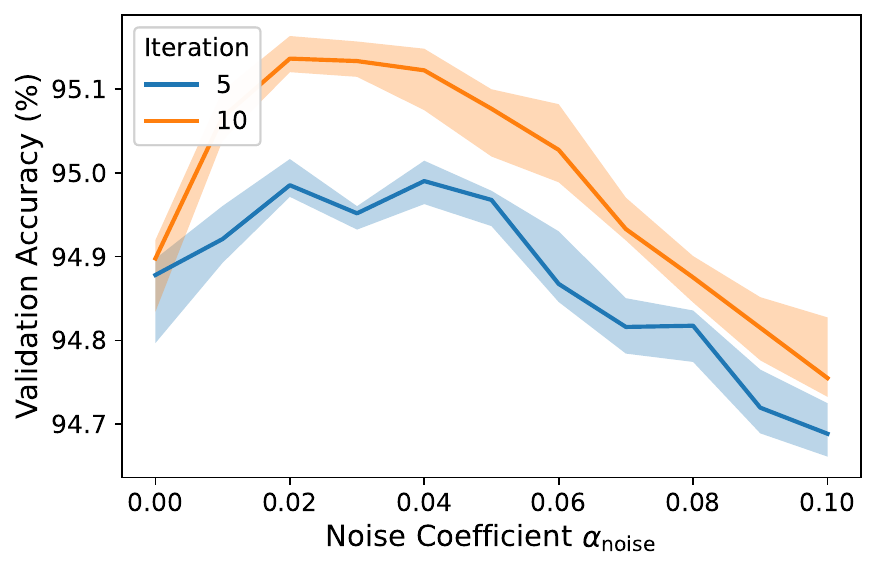} 
    \end{subfigure}
    \hspace{0.02\textwidth} 
    \begin{subfigure}[t]{0.47\textwidth}
        \centering
        \includegraphics[width=\linewidth]{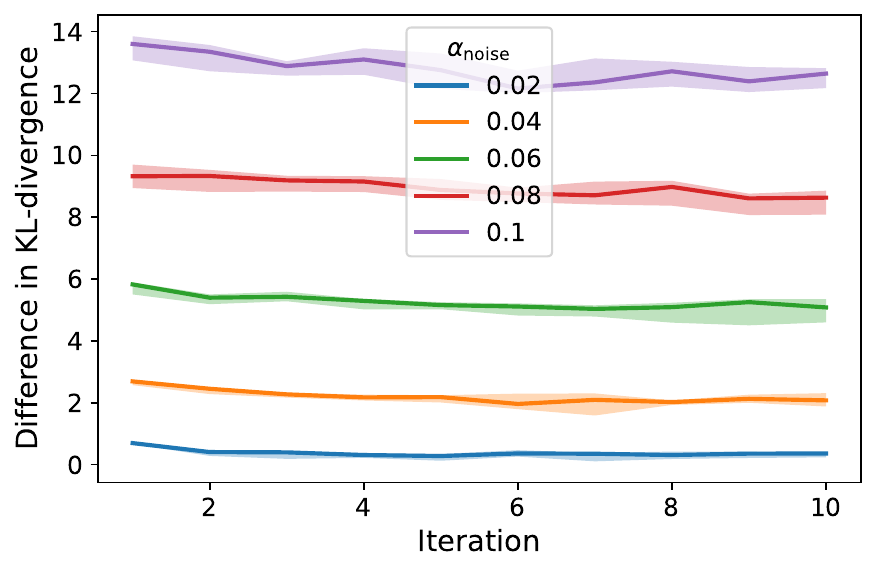} 
    \end{subfigure}
    \begin{subfigure}[t]{0.47\textwidth}
        \centering
        \includegraphics[width=\linewidth]{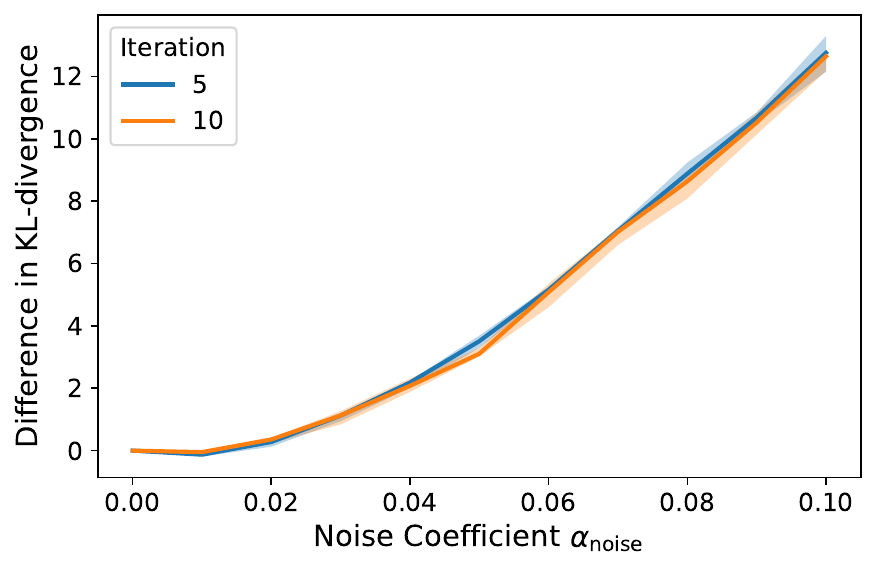} 
    \end{subfigure}
    \caption{
        \textbf{NAS-Bench-301: GEO with noisy tensor entries} 
        \textit{(Top Left):} GEO performance on \textsc{NAS-Bench-301} problem, with tensor entries corrupted by Gaussian noise with standard deviation $\alpha_\text{noise}$.
        Higher is better.
        Lines are medians and shaded regions are 1st and 3rd quartiles out of 40 runs.
        \textit{(Top Right and Bottom):} 
        Difference in KL-divergence between the noisy and noiseless ($\alpha_\text{noise}=0$) generative models.
        At each iteration a noiseless ($\alpha_\text{noise}=0$) generative model is trained on the same data as the noisy model used for GEO.
        KL-divergence is relative to the Boltzmann distribution from which the training data that iteration is drawn.
        Lines are medians and shaded regions are 1st and 3rd quartiles out of 10 independent runs.
        We see that KL-divergence is higher for the noisy model and that the difference in KL-divergence between noisy and noiseless models increases with increasing noise (increasing $\alpha_\text{noise}$).
    }
    \label{fig:nas_noisy_entries}
\end{figure}

With validation accuracy of the surrogate model as our objective function (see \ref{appendix:nas_obj_func}) we performed an experiment following the same procedure as in Section \ref{sec:noisy_entries}.
After training, random values independently drawn from a normal distribution with standard deviation $\alpha_\text{noise}$ are added to every tensor entry.
We recorded the performance of GEO on \textsc{NAS-Bench-301} with different values for the noise coefficient $\alpha_\text{noise}$.
We find that the optimal amount of noise is not zero.
Rather, several values of $\alpha_\text{noise}$ gave better performance, with $\alpha_\text{noise}=0.02$ and $\alpha_\text{noise}=0.03$ being the best of the values tried.
See the top left plot of Fig.~\ref{fig:nas_noisy_entries}.

As in Section \ref{sec:noisy_entries}, adding noise to the tensor entries of the model increases its KL-divergence relative to the distribution from which its training data is drawn, in this sense making it a worse model.
See the top right and bottom plots of Fig.~\ref{fig:noisy_entries}.
% We confirmed that this was indeed the case for the models used in GEO by calculating the KL-divergence of the trained generative model before and after the addition of noise and comparing the values.
Furthermore we found that in the context of a GEO run the generative models in runs with a higher $\alpha_\text{noise}$ have a higher difference in KL-divergence from the noise-free model.
We see yet again that the generative models that lead to the best performance of GEO are \emph{not} the ones best at modeling the distribution from which their training data is drawn.

\section{Benchmarking Tensor Network EDAs}
\label{sec:comparison} 

\begin{figure}
    \centering  
    \captionsetup{width=.9\linewidth}
    \includegraphics[width=1\linewidth]{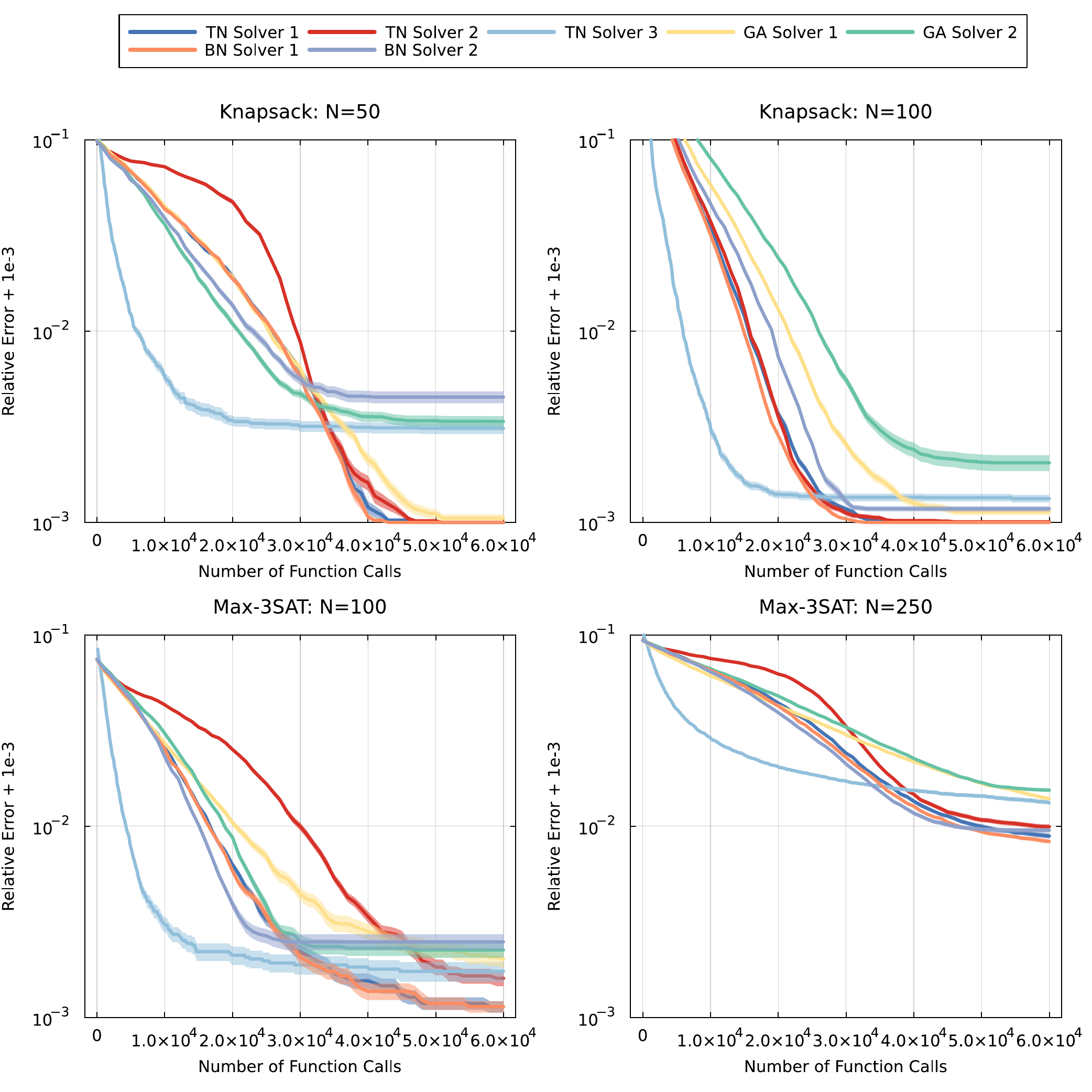}
    \caption{\textbf{Comparison of different evolutionary algorithms}. Relative errors of different solvers at solving \textsc{Knapsack} and \textsc{Max-3SAT} problem instances (see main text). The solvers stop when they use $6E4$ function calls. Error bars correspond to standard errors over 50 different runs of each solver. Our proposed solver \textsc{TN Solver 1}, and \textsc{BN Solver 1}, achieve lower relative errors.}
    \label{fig:benchmarks}
\end{figure}
In the previous section we saw that MPS generative models with mutation rate 0.01 and MPS models with lower bond dimension both performed relatively well as part of an EDA optimizer.
In this section we compare the performance of an optimizer with both an 0.01 bit flip mutation rate and a bond dimension 2 MPS against a selection of other evolutionary algorithms in a suite of combinatorial optimization problems. 
The following benchmark problems are chosen: Two \textsc{Knapsack} problems with $N=50$ and $N=100$ bits from \cite{dong2021phase} (listed there as $k_3$ and $k_5$, respectively), and Uniform Random \textsc{Max-3SAT} problems in the phase transition region from SATLIB \cite{maxsat}, with $N=100$ (and 430 clauses), $N=250$ (and 1065 clauses), number of variables. 
We use the 6th problem instance for each of these. 

We compare the following EDA solvers: 
our proposed TN solver consisting of Boltzmann selection with a Boltzmann distribution constructed from the top best 1000 samples from all previous generations and a mutation rate of value 0.01, denoted as \textsc{TN Solver 1}; 
GEO solver consisting of Boltzmann selection with a Boltzmann distribution constructed from unique samples from all previous generations and denoted as \textsc{TN Solver 2};
PROTES solver consisting of greedy selection and denoted as \textsc{TN Solver 3};
and two Bayesian networks with chain topology following the natural ordering, one with Boltzmann selection and mutation as in \textsc{TN Solver 1}, which we denote as \textsc{BN Solver 1}, and another with a 3-ary tournament selection and no mutation and denoted as \textsc{BN Solver 2}. 
The Bayesian networks are estimated at every generation using maximum likelihood estimation using the python library \textsc{pgmpy.py} \cite{Ankan2015}. 
Finally, we also include two versions of genetic algorithms with two-point crossover (with crossover rate 1): one with Boltzmann selection and mutation as in \textsc{TN Solver 1} which we denote as \textsc{GA Solver 1} and another with 3-ary tournament selection and no mutation which we denote as \textsc{GA Solver 2}. 
For all solvers using Boltzmann selection, the temperature is annealed as 
$T_t = T_0^{1 - t/t_{\rm max}}$,
where $t_{\rm max}$ is the total number of generations and the initial temperature $T_0$ is chosen to be the standard deviation of the (random) initial samples' costs. 
This choice guarantees that the final temperature corresponds to roughly the minimum possible gap between the optimal solution and the next best solution, which should be $\mathcal{O}(1)$ since the coefficients in all problems considered are integer-based. 
In all our experiments (with Boltzmann or otherwise) we limit the number of cost function calls to 60,000. 
In all evolutionary algorithms the population size is $1000$
(for EDA this corresponds to both the training samples as well as output samples of the models), 
except for \textsc{TN Solver 2} where we choose the default parameters from \cite{batsheva2024protes} with 10 training samples and 100 generated samples. 
(Some preliminary results with varying number of training/generated samples as well as bond dimension and learning rate of the TN model did not seem to impact the performance greatly, consistent with the observations of \cite{batsheva2024protes}). 
For \textsc{TN Solver 1} and \textsc{TN Solver 2} we choose a bond dimension of 2, a learning rate of $0.15$ and a single training step.
As in the portfolio optimization problem, we find empirically that the performance of the algorithm diminishes with increasing bond dimension.
It also diminishes with number of training steps (even for smaller learning rates).
The MPS models for these two solvers are trained from scratch at each generation.

Note that we do not make a comparison here against solvers specialized to the particular problems, i.e.\ solvers that exploit any special structure of \textsc{Knapsack} or \textsc{Max-3SAT}.
The intention is to evaluate the chosen solvers as ``black box" solvers.
This is the setting for which EDAs are most often intended and the setting in which, for example, \cite{original_geo} and \cite{batsheva2024protes} propose their algorithms.

Fig.~\ref{fig:benchmarks} shows the relative error of all solvers \textit{vs.} number of function calls with error bars corresponding to standard errors computed from 50 independent runs. 
For \textsc{Knapsack} both \textsc{BN Solver 1} and \textsc{TN Solver 1} find the optimal solution in all 50 instances in both problems. 
This is followed by \textsc{TN Solver 2} that solves $N=50$ all times and $N=100$ all but one time. 
For \textsc{Max-3SAT} we find that both \textsc{BN Solver 1} and \textsc{TN Solver 1} solve the $N=100$ instance in 47 out of 50 while \textsc{TN Solver 2} does in 37 out of 50. 
For $N=250$, none of the solvers is able to find the optimum in the allotted number of function calls. 
Nevertheless \textsc{BN Solver 1} and \textsc{TN Solver 1} approximate the optimum within 1\% on average. 
Our results indicate that \textsc{BN Solver 1} and our proposed solver \textsc{TN Solver 1} lead to the best results on these benchmark problems.

\section{Discussion}
\label{sec:discussion}

\subsection{Mutation Step}
Many optimization algorithms can be understood in terms of an exploit/explore dichotomy.
Genetic algorithms exhibit this balance explicitly as they, to some extent, separate the exploit and explore functions into two different operators, crossover and mutation.
Crossover suggests new solutions that are meant to resemble previously identified good solutions, while mutation leads to exploration of solutions with previously unseen features.
The form of genetic algorithms makes the balance of explore and exploit explicit through hyperparameters like the mutation rate or the steepness of selection for crossover.
This allows a user to tune the balance for optimal performance.

The experiments of Section \ref{sec:worse_models} suggest that the role of the generative model in GEO is not merely to learn features of good solutions and generalize to unseen good solutions.
This is evidenced by the fact that generative models that are strictly better at this task do not always lead to improved performance of GEO.
In addition to exploiting learned features of good solutions, the generative model is also performing an exploration function, one that is sometimes done better by models that are worse at learning the training data distribution.

In fact, at some level it is reasonable to expect a trade-off between these two jobs the generative model is being asked to perform.
At one extreme, a generative model whose output perfectly matches the distribution from which its training data is sampled will never output solutions better than those that have already been seen (because the training data are always sampled from already encountered solutions).
On the other hand, a generative model in some sense optimized for exploration might ignore potentially useful information learned from already encountered solutions, or worse might actively avoid features associated with good solutions.

This all suggests that a framework which separates these two jobs might provide better results than GEO.
We propose the addition of an explicit ``exploration" step occurring after the ``exploitation" step of training and sampling from the generative model.
Because it is simple to implement, we suggest random independent coordinate-wise mutations with tunable mutation rate parameter, though smarter mutations may, of course, be preferable for constrained search spaces.
In any case, with the exploitation and exploration tasks separated out into different components of the algorithm, improvements in the generative model have a greater chance of leading to an improvement in the optimization performance. 

The idea of adding an explicit mutation step in an EDA (like GEO) is not new. 
In \cite{handa2007effectiveness} the author showed that adding an explicit mutation operator improved the performance of various EDAs based on different choices of probabilistic graphical models, especially in the regimes of low population sizes. 
It would be interesting to further study the role of the population size on the one hand and the interplay of mutation and the quality of the generative model on the other.

\subsection{Comment on Generalization}
In the context of unsupervised generative modeling, generalization is the process of inferring a distribution from iid example data points \cite{zhao2018bias}.
Generalization is successful according to how similar the learned distribution is to the distribution from which the iid examples were taken.
This is the definition of generalization we have in mind throughout the paper and it is the metric on which we compare models in Section \ref{sec:worse_models}, declaring them ``better" or ``worse."
When using a generative model to generate samples, one may very well have other goals that don't amount to simply generating new samples that are identically distributed to the example data. 
We are somewhat reluctant to call these different goals generalization.
In these cases, it seems better to simply concede that not everything one might ask of a generative model is generalization than to stretch the concept to apply to conceptually unrelated tasks, and especially to tasks that could be in trade-off with generalization in the narrow sense.
In particular, tasks that involve sampling from a distribution from which one has no samples would not naturally be called generalization. 
This would include the task of generating samples with low values under some objective function, for example. 
This perspective is in contrast to that presented in \cite{generalization_metrics_geo}. 

This is not to say that performance at generalization (as we understand it here) is always easy to quantify.
On the contrary, it is often the case that there is neither a given, explicit distribution from which example data are iid samples
nor an endless supply of iid data points held out that might implicitly define such a distribution. 
In such situations, one often hopes to have \emph{enough} data to approximate a comparison to a distribution that is merely hypothetical, by calculating the likelihood or cross-entropy for held-out data, for example.
For high-dimensional distributions, these approximations based on held-out data are often inadequate.
These difficulties are merely difficulties in comparing the generated and true distributions, however.
Successful generalization is still understood to mean having similar such distributions.

\subsection{Conclusion}
The experiments in this work can be taken as an illustration of a more general principle: for evolutionary algorithms with generative ML components, increasing the quality of the generative model, all else held equal, does not necessarily lead to better performance of the algorithm as a whole.
The design of the evolutionary algorithm should be conscious of this and should provide compensating ``exploration" if need be.
This could come in the form of additional mutation as proposed above or by having the model be regularized more than would be necessary to simply avoid overfitting.
This regularization might be implemented by choosing a less expressive model (see Section \ref{sec:low_bond_dimension}), early stopping, or by adding an entropy term to the loss function used for training.
Other means of inducing additional exploration not explored in this work could include using distributions other than the Boltzmann distribution $p(x) \propto e^{-f(x)/T}$ to select training data from banked solutions.
For example one could use the distribution $p(x) \propto 1/(1 + e^{(f(x) - \nu)/T})$, which in essence chooses randomly from solutions below a soft cutoff $\nu$, and tune $\nu$ to allow solutions that wouldn't otherwise be competitive.
The added diversity in the training data could plausibly lead the trained generative model to output solutions that are more unlike solutions already seen.
None of these are guaranteed to work given a particular generative model, of course.
Overall it is unclear how to always make the most of an available powerful generative model.

It is worth pointing out that it is certainly possible for powerful models to be usefully used in evolutionary optimization routines.
See \cite{openai2022, deepmind2023, meyerson2023language} for examples where powerful LLMs allow optimization that likely wouldn't be possible otherwise.
We leave it as an open question how best to incorporate a better generative model into an evolutionary algorithm in a way where the additional power necessarily adds benefit.

Our results on the TN-EDA performance indicate that a low bond dimension MPS Born machine (of bond dimension 2 in all our experiments) yields better results than more expressible, higher bond dimension tensor networks. 
This is in contrast to optimization methods utilizing tensor networks to simulate quantum circuits, such as the quantum approximate optimization algorithm circuits \cite{dupont2022calibrating} or imaginary time evolution \cite{luchnikov2024largescalequantumannealingsimulation}, where lower bond dimension would be expected to result in worse optimizer performance. 

A low bond dimension MPS resulted in a TN-EDA comparable to a simple Bayesian Network EDA in some canonical combinatorial optimization problems. 
Higher bond dimensions did not lead to better results. 
This raises the question of whether, or in what regime, tensor networks offer any advantage compared to Bayesian networks within an EDA. 
A more promising regime for TN-EDA may be constrained combinatorial optimization. 
In particular, for some classes of global linear constraints, feasibility can be embedded directly into a tensor-network parameterization (e.g., via structured/block-sparse constructions), so that infeasible assignments have zero support by design \cite{symmetric_geo, linearly_constrained_geo}. 
Related constraint-aware constructions are possible in BN-EDAs as well (e.g., with auxiliary state variables), but can require more model-specific engineering. 
We therefore position TN-EDA not as universally superior to BN-EDA, but as a complementary approach that may offer advantages in constraint-structured regimes.

In this discussion we have suggested an explanation in terms of the exploration/exploitation trade-off for the observation that worse generative models can sometimes lead to better-performing EDA optimizers.
While this explanation is intuitive and reasonable on its face, it would be satisfying to see a setting in which this intuition could be formalized and made precise and where the hypothesis could be rigorously tested.
We consider this an interesting avenue for further work.

\section*{Acknowledgments}
The authors thank Douglas Hamilton for insightful discussions and support throughout the project.
The authors also thank Grecia Castelazo Martinez and Gordon Ma for help on an earlier version of this project.

\bibliographystyle{elsarticle-num}
\bibliography{tn_eda}

\appendix

\section{Details of Experiments in Section \ref{sec:worse_models}}
\label{appendix:port_opt_details}
In this appendix we provide methodological details of the experiments of Section \ref{sec:worse_models}.
Each experiment in that section consisted of several runs of GEO.
Within each generation of GEO the training data was 1000 solutions sampled from the bank of evaluated solutions according to a Boltzmann distribution.
After training the generative model on these, the generative model was sampled 1000 times and objective function values were evaluated for any samples not already in the bank of solutions.
All new solutions thus found were added to the bank before the next iteration of GEO.
All solutions evaluated at any point were retained in the bank for the duration of the run.
Before the first iteration, the solution bank was initialized by evaluating 100 random solutions chosen uniformly.

\subsection{The Portfolio Optimization Objective Function}
The objective function $f:\{0, 1\}^d\rightarrow\mathbb{R}$ optimized in the experiments of Section \ref{sec:port_opt} is
\begin{equation*}
f(x) = 
\begin{cases}
\frac{x^T\Sigma x}{\left(x^Tx\right)^2} & \text{if $n_\text{min} \leq x^Tx \leq n_\text{max}$,}\\
100(x^Tx - n_\text{max}) & x^Tx > n_\text{max}\\
100(n_\text{min} - x^Tx) & x^Tx < n_\text{max},
\end{cases}
\end{equation*}
where $n_\text{min}=20$ and $n_\text{max}=30$, and $\Sigma$ is the covariance matrix between the daily returns of the $d=107$ stocks present in Nasdaq's NDX index on 2017-12-15.
This was estimated from a year of historical closing price data from 2016-12-19 to 2017-12-15.
This represents a minimum-volatility, equal-weighted portfolio optimization with (soft) cardinality constraints.

The cardinality constraints are made soft to allow the algorithm to learn the constraints (rather than having them hard-coded into the representation).
The constant 100 here is simply chosen to be larger than the scale of $x^T\Sigma x/(x^T x)^2$.
In practice, the algorithm first learns to output portfolios that satisfy the cardinality constraint then reduces volatility within the region of valid portfolios.

\subsection{\textsc{NAS-Bench-301} Objective Function}
\label{appendix:nas_obj_func}
\textsc{NAS-Bench-301} provides a surrogate model that predicts the validation and test accuracy of a given convolutional neural net architecture on the CIFAR-10 benchmark.
Allowed architectures are made up of repeating sub-networks called ``cells."
There are two types of cells, Normal and Reduce, and all cells of the same type share the same architecture.
The search space of \textsc{NAS-Bench-301} is over the possible architectures for the Normal and Reduce cells.
The specifics of the allowed architectures are given in \cite{liu2019dartsdifferentiablearchitecturesearch}.
An architecture can be represented as a genotype, in practice a list of integers (16 for each cell type) that satisfies certain conditions.
See \cite{zela2022surrogatenasbenchmarksgoing} and the accompanying code for details.
Our objective function is defined using the SNB-DARTS-XGB v1.0 surrogate model.
We define it to return the validation accuracy output by the surrogate for valid genotypes and to return $\infty$ for invalid genotypes.

\subsection{MPS Training and Bond Dimension}
\label{appendix:training}
% All training of MPS tensor networks in this work was done with the procedure outlined in \cite{han2018unsupervised}.
% In a process similar to the 2-site DMRG algorithm, the first and second tensors in the network are updated in tandem using some number of gradient descent steps, then the second and third tensors are so updated, then the third and fourth, and so on down the line of tensors.
% Once the last pair of neighboring tensors is updated, the training proceeds back up the line to the first pair.
% So one round of training involves a sweep down and back along the line of tensors.
We use the training algorithm explained in \ref{appendix:training_algo} to train all MPS tensor networks in this work.
For the experiments in Section \ref{sec:worse_models} the gradient descent procedure for each pair of neighboring tensors consisted of a single gradient descent step with learning rate 0.1.
For experiments in Sections \ref{sec:bit_flip_noise} and \ref{sec:noisy_entries}, one training sweep down and back was performed.
For the experiment in Section \ref{sec:low_bond_dimension}, three training sweeps down and back were performed.
For the experiment in Section \ref{sec:nas_noisy_entries} five sweeps back and forth were performed.
At each iteration of the GEO algorithm we train the MPS starting from scratch, i.e.\ from a new random re-initialization.
This ensures the training data are all iid samples from a known distribution and allows us to easily characterize the model's quality by its KL-divergence relative that distribution.
We also found through informal experimentation that training the model from scratch each generation often improved performance of the GEO algorithm.
We speculate that that the random re-initializations provided additional beneficial exploration over the solution space.

As in \cite{han2018unsupervised}, the MPS bond dimensions in Sections \ref{sec:bit_flip_noise}, \ref{sec:noisy_entries}, and \ref{sec:nas_noisy_entries} were determined dynamically by enforcing a lower bound cutoff ratio, in our case $10^{-6}$, for singular values when canonicalizing after a gradient descent update.
In these experiments the maximum dynamical bond dimension was set to 5.
For the experiments of Section \ref{sec:low_bond_dimension} no cutoff was used and the bond dimensions were not chosen dynamically, but rather fixed to their various values.

\subsection{Tensor Network Hierarchical Clustering}
\label{appendix:ordering_details}
Several of our experiments (including all the experiments of Section \ref{sec:worse_models}) involve matrix product state (MPS) tensor networks, an architecture where the tensors are contracted in a line.
Each uncontracted leg along the MPS corresponds to a coordinate in the solution space (or in the context of probabilistic modeling corresponds to a random variable).
Because the structure of the MPS is not symmetric under a relabeling of the uncontracted legs, there is a meaningful choice to be made as to which legs correspond to which coordinates/variables.
Intuitively, placing variables that are more correlated closer to each other in the MPS structure would allow the MPS to better model those correlations for a given bond dimension.

To that end, in our experiments solving the equal-weighted portfolio optimization problem, we order the variables, in this case bits representing the presence or absence of assets, such that assets with highly correlated returns are nearer to each other in the MPS line structure.
We do this in two steps:
First, using pairwise hierarchical clustering (via the Ward variance minimization method) we obtain a linkage tree whose leaves are the assets that are under consideration.
As the distance measure we use
$
d(X, Y) = \frac{1}{\pi}\arccos{\operatorname{corr}(X, Y)},
$
where $\operatorname{corr}(X, Y)$ is the correlation between the daily returns of assets $X$ and $Y$.
Second, we order the assets to reduce distances between nearest neighbors while keeping clusters at every level of the tree contiguous. 
This was all implemented using the SciPy \verb|cluster.hierarchy| library.

The above ordering procedure was used for all the experiments in Section \ref{sec:port_opt}. 
For problems other than the equal-weighted portfolio optimization problem, we choose the ordering of the variables in the MPS to be the original ordering as found in the references from which those problems were drawn.

\subsection{GEO Temperature}
At each iteration of GEO solutions are selected from a Boltzmann distribution over a bank of already evaluated solutions.
Correctly choosing the temperature parameter of the Boltzmann distribution is crucial for algorithm performance.
Choose it too high and convergence is slow as the generative model is trained on (and thus outputs) lower quality solutions.
Choose it too low and the generative model will be trained on too few solutions, such that its output will fail to adequately explore the solution space, leading to the algorithm becoming stuck at local minima.
Based on these considerations, for the experiments of Section \ref{sec:worse_models}, we chose the temperature such that the fifth best solution is one third as likely as the best solution, i.e.\ $T$ such that $e^{f(x_5)/T} = e^{f(x_1)/T}/3$ where $x_5$ is the fifth best solution in the bank and $x_1$ is the best solution in the bank, so $T=\big(f(x_5) - f(x_1)\big)/\log{3}$.
This heuristic ensures that the training dataset never becomes overwhelmingly dominated by just one solution no matter how much better it is than the other solutions in the bank.
The values 5 and 3.0 were determined to work well on similar problems through informal experimentation, and in practice they would need to be tuned for any particular problem.
In situations where the there is a five-way or more tie for best we set the temperature to be $T=\big(f(x') - f(x_1)\big)/\log{3}$ where $x'$ is the best solution not involved in the tie for first, to ensure we do not end up with a temperature of 0.

For the other experiments  (i.e.\ those of Section \ref{sec:comparison}) the temperature used for the Boltzmann selection steps is chosen to be $T_t = T_0^{1 - t/t_{\rm max}}$ for simplicity, rather than using the above adaptive algorithm.

\section{KL-divergence of Noisy Tensor Network Generative Models}
\label{appendix:kl_divergence}

\begin{figure}
    \centering
    \captionsetup{width=.9\linewidth}
    \begin{subfigure}[t]{0.47\textwidth}\label{undiffused_model}
        \centering
        \includegraphics[width=\linewidth]{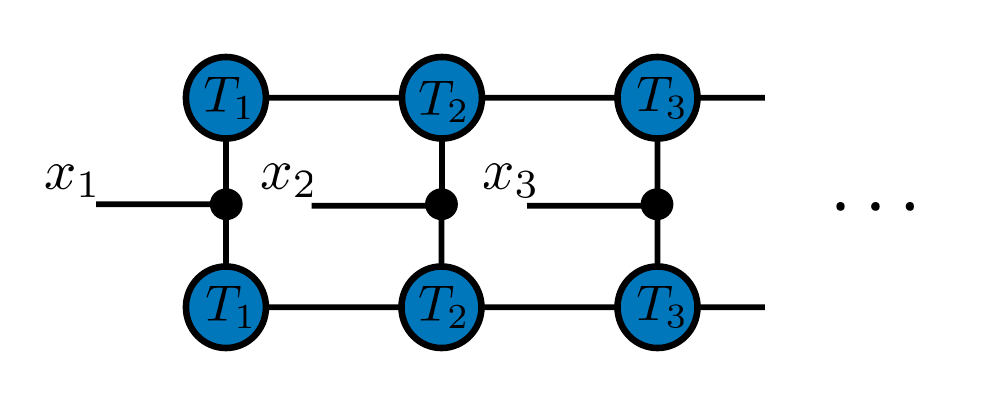} 
    \end{subfigure}
    \hspace{0.02\textwidth} 
    \begin{subfigure}[t]{0.47\textwidth}\label{diffused_model}
        \centering
        \includegraphics[width=\linewidth]{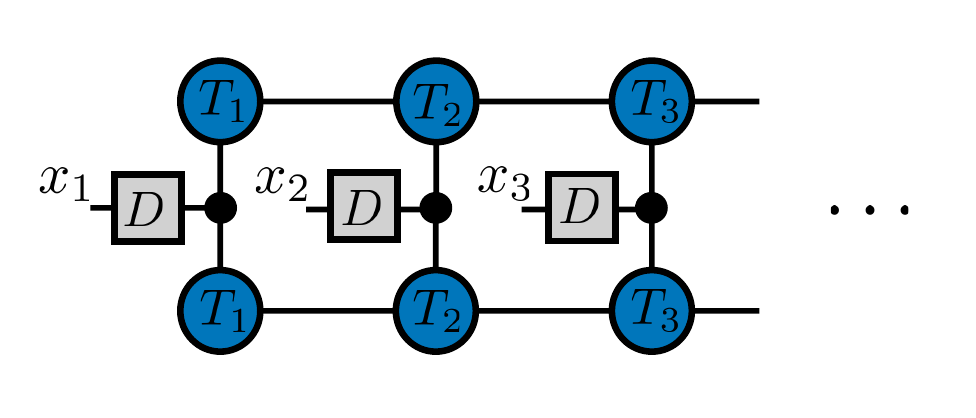} 
    \end{subfigure}
    \caption{
        \textbf{Effect of bit flips}.
        The probabilities of bit strings $x_1x_2\cdots x_d$ in the original model are given by the tensor network on the left.
        Following sample generation from this model by independent random bit flips effectively gives a new generative model, a diffused version of the original, which can be obtained by acting on the legs with a diffusion operator 
        $D = \begin{bmatrix}
            1 - p_\text{flip} & p_\text{flip} \\
            p_\text{flip} & 1 - p_\text{flip}
        \end{bmatrix}$
        where $p_\text{flip}$ is the probability of a bit flipping.
    }
    \label{fig:undiffused_vs_diffused}
\end{figure}

\begin{figure}
    \centering
    \captionsetup{width=.9\linewidth}
    \begin{subfigure}[t]{0.5\textwidth}\label{contraction_diffused}
        \centering
        \includegraphics[width=\linewidth]{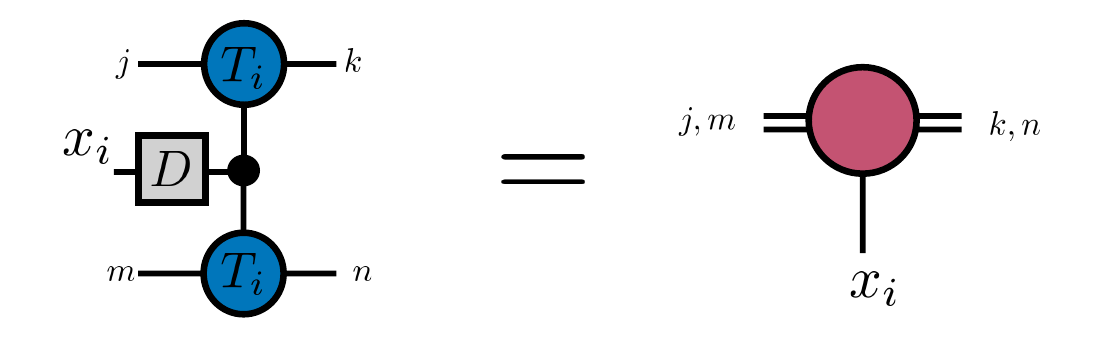} 
    \end{subfigure}
    \caption{
        \textbf{Efficient evaluation of diffused probabilities}.
        Contracting tensors as above allows for efficient evaluation of the probability of any bit string under the diffused model, i.e.\ the model followed by random independent bit flips.
    }
    \label{fig:contraction_diffused.pdf}
\end{figure}

In Section \ref{sec:worse_models}, to demonstrate that models followed by bit flips, models with noise added to the tensor entries, or models with lower bond dimension all perform worse at learning the probability distribution from which their training data was drawn, we compare KL-divergences of the model outputs relative to this distribution.
Because the bank of solutions from which training data is drawn is finite, we can calculate these KL-divergences exactly by simply summing over finitely many terms.
This is straightforward for the MPS models with different bond dimension or with noise added to their tensor entries, as calculating probabilities can be done efficiently given an MPS.
For the case of the MPS considered in Section \ref{sec:bit_flip_noise}, whose sampled bit strings are subsequently subject to random bit flips, the exact KL-divergence can still be calculated efficiently, but with an additional step.
First note that sampling followed by bit-flip noise, taken together, is equivalent to sampling from a diffused version of the original model:
Let $q(x_1\cdots x_d)$ be the probability of sampling a given bit string $x_1\cdots x_d$ from the original model, and let $\tilde{q}(x_1\cdots x_d)$ be the probability of obtaining $x_1\cdots x_d$ by the process of sampling followed by independent random bit flips.
Then, because the bit flips are independent, we have
\begin{equation}\label{bitflip_equation}
\begin{aligned}
    \tilde{q}(x) &= \sum_y p(x|y)q(y) \\
    &= \sum_{y_1\cdots y_d} p(x_1|y_1)p(x_2|y_2)\cdots p(x_d|y_d) q(y_1\cdots y_d),
\end{aligned}
\end{equation}
where $p(x_i|y_i)$ is the probability that an original bit $y_i$ becomes a corrupted bit $x_i$. 
We see that, given a tensor network form for the original probabilities $q(y_1\cdots y_d)$, one only needs to act on the uncontracted legs with the matrices $p(x_i|y_i)$ which in our case are simply
\[
D 
= 
\begin{bmatrix}
    p(0|0) & p(0|1) \\
    p(1|0) & p(1|1)
\end{bmatrix}
=
\begin{bmatrix}
    1 - p_\text{flip} & p_\text{flip} \\
    p_\text{flip} & 1 - p_\text{flip}
\end{bmatrix}.
\]
Recall that the probability of a given bit string in the original MPS without bit-flip noise is given by the contraction diagram on the left of Figure \ref{fig:undiffused_vs_diffused}.
(Note, the small black dots in that diagram represent the diagonal tensor $\delta_{ijk}$, which is 1 when the value on all indices matches and zero otherwise.)
To obtain the probabilities of the model ``diffused" by bit-flip noise we contract the matrix $D$ to each uncontracted leg, as shown in the right of Figure \ref{fig:undiffused_vs_diffused}.
This can be efficiently evaluated by first performing the contractions shown in Figure \ref{fig:contraction_diffused.pdf}.
The result is then a new MPS that can be efficiently contracted.
To summarize, because we have $q(y_1\cdots y_d)$ in matrix-factorized form, the sum in \eqref{bitflip_equation} factorizes, removing the need to sum over exponentially many terms.

\section{Tensor Network Born Machine Training Algorithm}
\label{appendix:training_algo}

\begin{figure}
    \centering
    \includegraphics[width=0.5\linewidth]{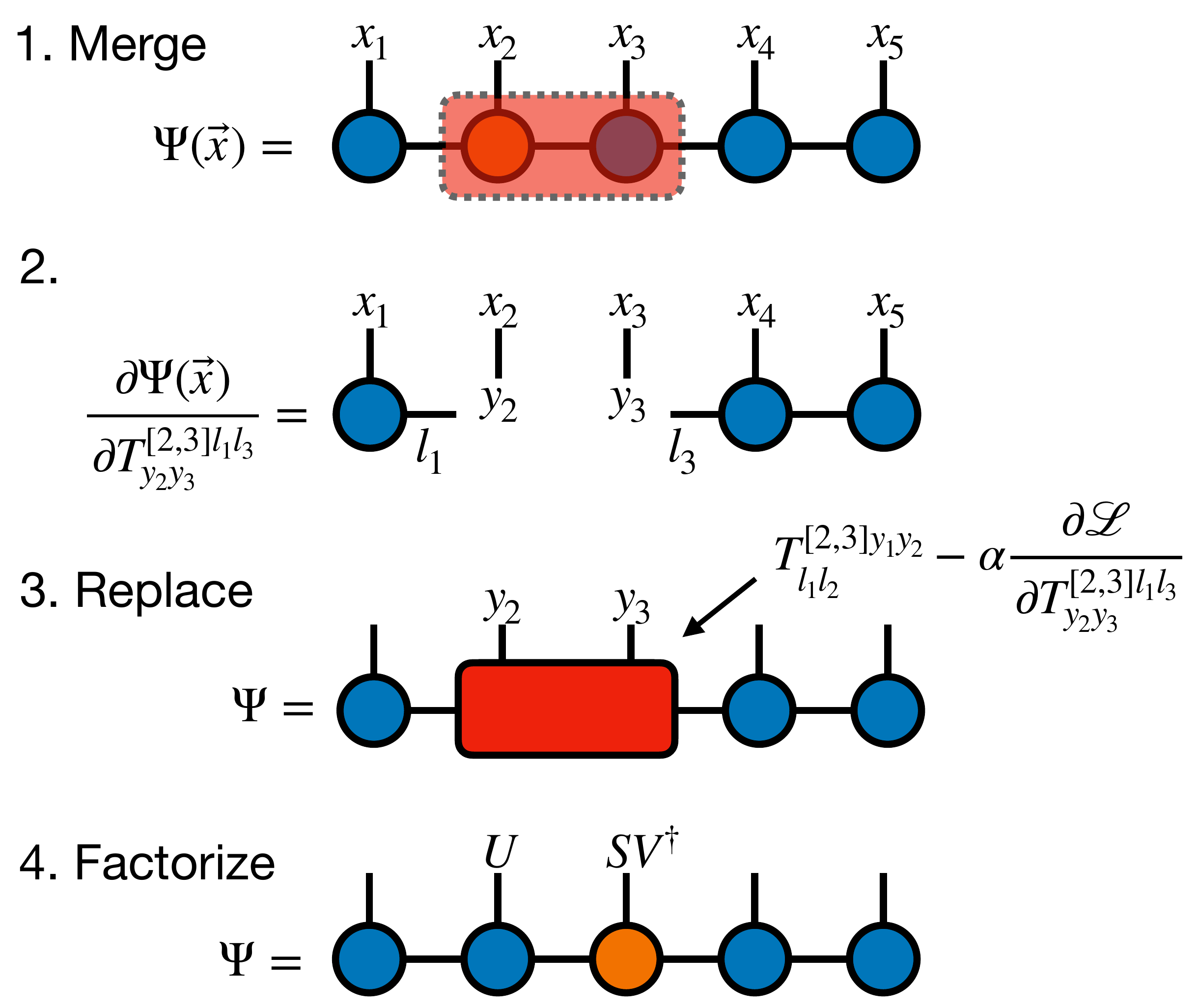}
    \caption{\textbf{Training steps for TNBM.} This series of steps is repeated from left-to-right and back.}
    \label{fig:training_tnbm}
\end{figure}

In this work we train the TNBM using the gradient descent algorithm described in Ref.~\cite{han2018unsupervised}. 
The loss function is given by the negative log likelihood: 
\begin{equation}\label{eq:loss}
    \mathcal{L}=-\frac{1}{\left|\mathcal{T}\right|}\sum_{\vec{x}}\log P(\vec{x}),    
\end{equation}
where $\mathcal{T}$ is the training set, $P(\vec{x})=|\Psi(\vec{x})|^2/Z$ is the Born probability represented by the TNBM (see Sec.~\ref{sec:born_bayes} for a description of the TNBM). 
To minimize the loss (\ref{eq:loss}) Ref.~\cite{han2018unsupervised} proposed to use an algorithm inspired by the DMRG method \cite{white1992density, schollwock2011density}. 
In Fig.~\ref{fig:training_tnbm} we present the main steps of the algorithm. 
The idea is to traverse the MPS representation of the TNBM from left to right, performing gradient updates with respect to two neighboring tensors at a time. 
At each step, the canonical tensor -- an MPS tensor with left (right) isometries on its left (right) side -- is shifted by one site. 
This ensures that the MPS remains in canonical form throughout training. 

The training procedure can be summarized in the following steps (illustrated in Fig.~\ref{fig:training_tnbm}):
\begin{enumerate}
        \item Merge: Combine two neighboring tensors into a single effective tensor block.
        \item Differentiate: Compute the gradient of the function $\Psi(\vec{x})$ with respect to the entries of the merged tensor.
        \item Update: Replace the merged tensor by performing a gradient step: $T \longrightarrow  T - \alpha \frac{\partial \mathcal{L}}{\partial T}$, where $\alpha$ is the learning rate.
        \item Factorize: Decompose the updated tensor back into two neighboring tensors via SVD, restoring the MPS structure and shifting the canonical center.
\end{enumerate}

These steps are repeated iteratively, sweeping across the MPS. 
Figure ~\ref{fig:training_tnbm} shows the schematic representation of these steps, where orange tensors denote the canonical center, and the sweeping direction is indicated by the shifting of the updated tensors.

% \section{TN-EDA for Neural Architecture Search}
% In this section we present results of our TN-EDA when compared to the following baselines: random search, genetic algorithm, and a one dimensional Bayesian network EDA. The motivation for this analysis is two-fold: on the one hand we want to give further evidence that more complex TNBMs (bigger bond dimension) lead to worse performance on a more realistic, challenging task (compared to the portfolio optimization instance considered in the main text). Second, we would like to see whether our TN-EDA is capable of outperforming the other baselines (our results from Section ? show that a simple Bayesian network EDA performs similar to our proposed TN-EDA). 

% We will use the \textsc{NAS-Bench-301} benchmark from \cite{zela2022surrogatenasbenchmarksgoing}. 
\end{document}